\documentclass[11pt]{article}

\usepackage[final]{acl}

\usepackage{times}
\usepackage{latexsym}

\usepackage[T1]{fontenc}

\usepackage[utf8]{inputenc}

\usepackage{microtype}

\usepackage{inconsolata}
\usepackage{booktabs}
\usepackage{graphicx}
\usepackage{adjustbox} 
\usepackage{amsmath}
\usepackage{multirow} 
\usepackage{enumitem}
\usepackage{tabularx}
\usepackage{float}
\usepackage{makecell}

\newcommand{\LCQA}{\textsc{Logical\mbox{-}CommonsenseQA}}

\title{\LCQA:\\ A Benchmark for Logical Commonsense Reasoning}

\author{Obed Junias \\
  University of Colorado Boulder\\
  \texttt{obed.junias@colorado.edu} \\\And
  Maria Leonor Pacheco \\
  University of Colorado Boulder\\
  \texttt{maria.pacheco@colorado.edu} \\}

\begin{document}
\maketitle
\begin{abstract}
Commonsense reasoning often involves evaluating multiple plausible interpretations rather than selecting a single atomic answer, yet most benchmarks rely on single-answer evaluation, obscuring whether statements are jointly plausible, mutually exclusive, or jointly implausible. We introduce \LCQA, a benchmark that reframes commonsense reasoning as \emph{logical composition} over pairs of atomic statements using plausibility-level operators (\textsc{AND}, \textsc{OR}, \textsc{NEITHER/NOR}). Evaluating instruction-tuned, reasoning-specialized, and fine-tuned models under zero-shot, few-shot, and chain-of-thought prompting, we find that while models perform reasonably on conjunctive and moderately on disjunctive reasoning, performance degrades sharply on negation-based questions. \LCQA~exposes fundamental reasoning limitations and provides a controlled framework for advancing compositional commonsense reasoning. 
\end{abstract}

\section{Introduction}

Commonsense reasoning is central to human cognition and a long-standing challenge in artificial intelligence and natural language understanding. Developmental studies suggest that for humans, it emerges from intuitive, experience-based frameworks in early childhood and becomes more abstract and socially grounded through embodied experience, social interaction, cultural transmission, and language~\cite{spelke1992origins, spelke2007core, gopnik2004finding, meltzoff2009foundations}. In contrast, large language models (LLMs) acquire ``commonsense-like'' knowledge indirectly from large-scale text corpora via objectives such as next-token prediction, rather than through grounded interaction or social learning~\cite{bender-koller-2020-climbing}. Despite this, LLMs exhibit emergent behaviors resembling human intuitive reasoning, including inferring intentions and anticipating socially plausible actions~\cite{sap2019atomic, bisk2020piqa}. However, this competence is brittle, as models fail under minimal semantic perturbations, struggle with logical consistency, and rely on statistical shortcuts rather than genuine reasoning~\cite{ettinger2020bert, geiger2020neural, mccoy-etal-2019-right, niven-kao-2019-probing}. To improve the evaluation of commonsense reasoning, we introduce \LCQA, a benchmark that reframes the task as a \textit{logical composition over multiple plausible answers}, requiring models to reason about relationships between plausible statements rather than select a single response.

Early approaches to commonsense reasoning relied on symbolic knowledge bases~\cite{lenat1995cyc,liu2004conceptnet}, which were expressive but limited in scalability and coverage. More recent benchmarks shift the evaluation towards the ability of neural models to infer plausible answers from text~\cite{talmor2019commonsenseqa,sap2019socialiqa,sap2019atomic,zellers-etal-2019-hellaswag,bisk2020piqa,sakaguchi2020winogrande}. However, these benchmarks reduce commonsense reasoning to single-answer prediction, despite the fact that many real-world situations admit multiple plausible interpretations~\cite{min-etal-2020-ambigqa}. For example, in the question \textit{``The fox walked from the city into the forest—what was it looking for?''} both \textit{``natural habitat''} and \textit{``shelter from disturbances''} are plausible answers. Restricting the evaluation to a single atomic choice removes this natural ambiguity and obscures whether models recognize jointly plausible, mutually exclusive, or jointly implausible interpretations. 
Beyond linguistic ambiguity, commonsense is both socially grounded and compositional. People judge commonsense statements not only by personal belief but also by estimating whether others would share that belief~\cite{whiting2024commonsense}, implying that commonsense reflects a distribution over plausible interpretations rather than a single canonical truth. At the same time, human reasoning integrates multiple alternatives through compositional inference, recognizing when statements are jointly plausible (\textsc{AND}), partially plausible (\textsc{OR}), or jointly implausible (\textsc{NEITHER/NOR}). These relational judgments are central to everyday reasoning about intentions, plans, and social norms, yet current benchmarks fail to evaluate whether models can reason over such socially grounded and compositional plausibility relationships.

In \LCQA, each instance pairs two atomic answers under one of three symbolic relations—\textsc{AND}, \textsc{OR}, or \textsc{NEITHER/NOR}. These are not strict propositional operators, but \emph{plausibility-level composition constructs} indicating joint, partial plausibility, or joint implausibility. This formulation preserves the multiple-choice question (MCQ) format of modern benchmarks while explicitly modeling ambiguity and compositional reasoning. We evaluate instruction-tuned, reasoning-specialized, and fine-tuned LLMs under zero-shot, few-shot, and chain-of-thought prompting. Results show that while models perform reasonably well on conjunctive and moderately on disjunctive reasoning, performance degrades sharply on negation-based compositions, revealing persistent gaps in the commonsense reasoning capabilities of state-of-the-art systems.

The dataset\footnote{\scriptsize{\url{https://huggingface.co/datasets/ojayy/logical-csqa}}}
and code\footnote{\scriptsize{\url{https://github.com/obedjunias19/logical-csqa}}} are publicly available.
\section{Related Work}

Commonsense reasoning has been widely studied through symbolic knowledge bases, neural models, and benchmark-driven evaluation. Early symbolic approaches such as \textsc{Cyc}~\cite{lenat1995cyc} and \textsc{ConceptNet}~\cite{liu2004conceptnet} enabled structured inference but suffered from limited coverage and scalability. More recent benchmarks, including \textsc{CommonsenseQA}~\cite{talmor2019commonsenseqa}, \textsc{SocialIQA}~\cite{sap2019socialiqa}, \textsc{ATOMIC}~\cite{sap2019atomic}, \textsc{PIQA}~\cite{bisk2020piqa}, \textsc{HellaSWAG}~\cite{zellers-etal-2019-hellaswag}, and \textsc{WinoGrande}~\cite{sakaguchi2020winogrande}, shift evaluation towards the ability of neural models to infer plausible responses from text. However, these benchmarks adopt a single atomic-answer format that fails to capture the inherent ambiguity of many real-world commonsense questions.

Prior work has emphasized that ambiguity and human disagreement are intrinsic to commonsense reasoning. \textsc{AmbigQA}~\cite{min-etal-2020-ambigqa} and \textsc{ProtoQA}~\cite{boratko2020protoqa} show that many questions naturally admit multiple valid answers, motivating the need to evaluate systems beyond single gold answers. Separately, logical reasoning benchmarks such as \textsc{LogiQA}~\cite{liu2020logiqa}, \textsc{ReClor}~\cite{yu2020reclor}, and \textsc{COM2}~\cite{fang2024complex} evaluate deductive inference using formal logical operators, but target logical validity rather than commonsense plausibility. Complementary work in the social sciences argues that commonsense is socially grounded, reflecting shared expectations rather than objective truth~\cite{whiting2024commonsense}.

Our work differs from these lines of research by introducing \LCQA, a benchmark that evaluates commonsense reasoning as \emph{compositional plausibility}. Instead of open-ended multi-answer annotation or formal deductive logic, \LCQA~preserves the MCQ format of modern commonsense benchmarks while explicitly encoding joint plausibility, partial plausibility, and joint implausibility through symbolic composition over human-validated atomic answers.
\begin{table*}[t]
\centering
\scriptsize
\begin{tabularx}{\textwidth}{p{1.8cm} p{4cm} X}
\toprule
\textbf{Operator} & \textbf{Interpretation} & \textbf{Example} \\
\midrule

$\boldsymbol{a \; \mathtt{AND} \; b}$ 
& Both statements independently plausible 
& \textit{Q: Sammy wanted to go to where the people were. Where might he go?}\\
& & A: local events AND social venues \\[6pt]

$\boldsymbol{a \; \mathtt{OR} \; b}$ 
& At least one statement plausible 
& \textit{Q: Sammy wanted to go to where the people were. Where might he go?}\\
& & A: local events OR empty parks \\[6pt]

$\boldsymbol{\mathtt{NEITHER} \; a \; \mathtt{NOR} \; b}$ 
& Neither statement plausible 
& \textit{Q: Sammy wanted to go to where the people were. Where might he go?}\\
& & A: NEITHER quiet retreats NOR empty parks \\

\bottomrule
\end{tabularx}
\caption{Composition operators used in \LCQA\ to express plausibility relations. 
}
\label{tab:logic_labels}
\end{table*}

\section{The Logical Commonsense Dataset}
\label{sec:commonsenseqa_logic}

We construct \LCQA~to evaluate compositional, multi-answer commonsense reasoning by extending \textsc{CommonsenseQA}~\cite{talmor2019commonsenseqa}, a benchmark of 12,247 MCQs with one correct answer and four distractors that probes commonsense relations such as cause–effect, purpose, and spatial association. \LCQA~reformulates these single-answer questions into logically composed items that explicitly model relationships among plausible alternatives while preserving full compatibility with the MCQ format.

\paragraph{Construction Pipeline} We build the dataset using a three-stage pipeline that combines neural generation with deterministic symbolic composition to convert atomic answers into logically structured multi-answer instances.

\textit{Stage 1: Generation of Candidate Options.} Starting from \textsc{CommonsenseQA} questions and their gold answers, we sample 5,000 instances and use GPT-4o-mini to over-generate a diverse set of atomic answer candidates. The model is prompted to produce both plausible and implausible alternatives, with an emphasis on multi-step causal or situational reasoning rather than shallow lexical cues. This process yields 4--6 plausible and 4--6 implausible candidates per question, spanning physical, social, and situational commonsense domains.

\textit{Stage 2: Refinement and Pruning.} We then use GPT-4o-mini to refine and filter the generated candidates according to three criteria: (1) removing logically inconsistent or factually incorrect answers; (2) eliminating trivial options that can be resolved through keyword matching; and (3) identifying highly plausible options that satisfy most semantic and contextual constraints of the question but fail due to a non-obvious commonsense violation. This stage yields a curated set of three correct and four incorrect atomic options per question, each requiring multi-step inference. After filtering, one sampled question is removed, leaving 4,999 curated question sets.

\textit{Stage 3: Deterministic Logical Composition.}
In the final stage, we use a symbolic program to deterministically combine pairs of refined atomic options into logical compositions labeled with one of the three base plausibility operators shown in Table~\ref{tab:logic_labels}. These relations are not strict propositional operators; rather, they encode commonsense judgments about joint plausibility. This stage yields 14,997 operator-specific instances in total, with 4,999 instances for each base operator. We additionally construct a \textsc{Mixed} setting, in which different operators appear across answer choices within the same question, yielding 4,999 additional instances.

Prompt templates for all stages of the construction pipeline are provided in Appendix~\ref{app:construction_prompts}.

\paragraph{Human Validation} To ensure the commonsense validity of the refined atomic options, we conducted a human validation study using the awareness--consensus framework proposed by~\citet{whiting2024commonsense}.
Two independent annotators evaluated 250 questions from the Stage~2 test split, corresponding to 50\% of the 500-question test set. For each question, annotators assessed the refined set of three correct and four incorrect atomic options. For each atomic option, annotators assessed (1) whether they personally believed the option was correct (awareness) and (2) whether they believed most others would agree (consensus). 
This dual assessment captures both individual plausibility and perceived social agreement, aligning with recent views of commonsense as socially grounded. Inter-annotator agreement, measured using Gwet's AC$_2$ with ordinal weighting~\cite{gwet2014handbook}, indicated high agreement for both awareness (AC$_2$ = 0.84) and consensus (AC$_2$ = 0.91), suggesting that the refined options are both semantically coherent and socially interpretable. Disagreements between annotators were then resolved through adjudication.

We report label accuracy on the adjudicated dataset after composition into \textsc{AND}, \textsc{OR}, \textsc{NEITHER/NOR}, and \textsc{Mixed} instances. Performance is high for \textsc{AND} (89.2\%), \textsc{OR} (96.4\%), and \textsc{Mixed} (88.4\%), and reasonable for \textsc{NEITHER/NOR} (73.6\%). These numbers reflect a reasonably accurate labeling of composed instances.

\paragraph{Resulting Dataset}

After Stage 2, we obtain 4,999 curated questions with refined atomic answer sets. In Stage 3, each question is expanded into logically structured instances evenly distributed across the three base operators, with an additional \textsc{Mixed} setting in which operators are randomly assigned across options. The final dataset contains 19,996 instances, with \textsc{AND}, \textsc{OR}, \textsc{NEITHER/NOR}, and \textsc{Mixed} each contributing 4,999 instances. 


\paragraph{Diversity Analysis}
Because atomic options are generated using an LLM, we explicitly evaluate lexical and semantic diversity to ensure that the dataset does not suffer from generation collapse.

Table~\ref{tab:diversity_table} reports lexical and semantic diversity statistics for atomic and composite options. Following \citet{li2016diversity}, we measure Distinct-2 (ratio of unique bigrams to total bigrams), obtaining 0.70 for atomic options and 0.23 for composite options, indicating substantial lexical variation in the generated atomic options. We further compute semantic diversity as the negation of mean pairwise cosine similarity between sentence embeddings~\cite{tevet2021evaluating}. Global semantic diversity is high for both atomic and composite options (0.89), suggesting low global semantic redundancy across the dataset, while intra-question diversity is 0.73 for atomic options and 0.47 for composite options, indicating substantial variation within each question's answer set.
\begin{table}[t]
\centering
\scriptsize
\begin{tabular}{lcc}
\toprule
\textbf{Metric} & \textbf{Atomic} & \textbf{Composite} \\
\midrule
Distinct-2 & 0.70 & 0.23 \\
Global Semantic Diversity & 0.89 & 0.89 \\
Intra-question Diversity & 0.73 & 0.47 \\
\bottomrule
\end{tabular}
\caption{Lexical and semantic diversity metrics for atomic and composite options.}
\label{tab:diversity_table}
\end{table}

Lower lexical and intra-question diversity for composite options is expected, as compositions deterministically reuse atomic statements and fixed operator tokens across options for the same question. Additionally, diversity is not left to unconstrained sampling: our Stage~1 generation explicitly enforces structured diversity through near-miss distractors, partial-truth alternatives, and mechanism-based variation, while prohibiting trivial or redundant options. 

Overall, these results suggest that the dataset exhibits controlled diversity arising from structured recombination and constraint-based generation. 

\section{Task Formulation}

Instances in \LCQA~consist of a natural language question and four answer options. Unlike standard benchmarks where each option is an atomic answer, each option in our dataset is a \emph{logical composition} of two independent atomic statements combined using one of three plausibility-level operators: \textsc{AND}, \textsc{OR}, or \textsc{NEITHER/NOR} (see Table ~\ref{tab:logic_labels}). 

Given a question, the model must select the option whose composite plausibility best matches the implied commonsense constraints. While the task remains a simple MCQ problem, it requires models to assess the plausibility of individual statements and reason about their interaction under the specified operator. We include both operator-specific conditions, in which all answer options for a question share the same logical operator, and a \textsc{Mixed} condition, in which different operators appear across options. While the operator-specific conditions allow for controlled evaluation of each relation type, the \textsc{Mixed} condition prevents models from exploiting operator-specific patterns and requires direct inference over the composed statements. Overall, this formulation preserves the MCQ format while substantially increasing the demands on compositional commonsense reasoning.

\section{Experiments} \label{sec:experiments}

We evaluate a diverse set of encoder-only classifiers, encoder--decoder models, decoder-only open-weight LLMs, and proprietary closed-source models on \LCQA\/. Open-weight decoder-only models are evaluated under zero-shot, few-shot, and supervised fine-tuning regimes, while proprietary closed-source models are evaluated in the zero-shot setting only. Full details on model selection, prompting strategies, and training procedures are provided in Appendices~\ref{app:models}, \ref{app:prompting}, and \ref{app:training}.

Following the dataset construction described in Section~\ref{sec:commonsenseqa_logic}, we use all 19,996 instances of \LCQA\, split into 11,996 training, 6,000 development, and 2,000 test examples. Splits are stratified by logical relation types, ensuring an even 25\% distribution across \textsc{AND}, \textsc{OR}, \textsc{NEITHER/NOR} operators, and the \textsc{Mixed} setting. We report accuracy and macro-F1 as evaluation metrics.

For analysis, we further divide the test set into a human-validated (HV) subset and a non-validated (NV) subset, each comprising 50\% of the test data. Table~\ref{tab:main_summary} summarizes the main results. 

Across all open-source model families, two consistent patterns emerge: (1) all models perform reasonably well on \textsc{AND} and moderately on \textsc{OR}, and (2) performance collapses on \textsc{NEITHER/NOR} in zero- and few-shot settings, even for the strongest LLMs. Observed weaknesses are amplified in the \textsc{Mixed} condition, where different operators appear across options and models must implicitly infer the governing relation. F1 drops to 43--56\% for all few-shot models, suggesting that these models do not reliably maintain operator-level representations and instead revert to surface-level heuristics. We observe the same qualitative trend for proprietary closed-source models evaluated in the zero-shot setting: they perform strongly on \textsc{AND}, moderately on \textsc{OR} and \textsc{Mixed}, and substantially worse on \textsc{NEITHER/NOR}. This consistency across open-weight and closed-source models suggests that the difficulty exposed by \LCQA\ is not specific to a particular model family, but reflects a broader challenge in composing plausibility under negation-based operators. In contrast, fine-tuned models achieve 83--95\% F1 across all operators, suggesting that the task is learnable with supervision and that zero- and few-shot failures, particularly on negation, reflect inference-time limitations rather than dataset artifacts.

To further contextualize these results, we also report a small-scale human evaluation on the human-validated subset, using 10 examples per condition. Human accuracy is 0.90 on \textsc{AND}, 0.70 on \textsc{OR}, 0.70 on \textsc{NEITHER/NOR}, and 0.90 on \textsc{Mixed}. Although limited in scale, these results suggest that human performance does not exhibit the same collapse on \textsc{NEITHER/NOR} observed for zero-shot LLMs, supporting the view that the benchmark remains solvable by humans while exposing systematic model weaknesses.

To better isolate the source of these failures, we perform a decomposition analysis that separates atomic plausibility estimation, operator grounding, and distractor competition. Results show that atomic plausibility alone does not explain the observed difficulty: on the human-validated subset, atomic plausibility classification accuracy is reasonably strong (79\%), but operator verification remains imperfect even with gold atomic labels (52\%--69\%), and accuracy degrades further once distractor composite options are introduced, with the largest drop observed for \textsc{NEITHER/NOR} (71\% to 47\%). This suggests that the challenge of \LCQA\ arises from the interaction of plausibility estimation, operator grounding, and distractor competition in the full multiple-choice setting.

Additional results—including varying numbers of shots, prompting strategies, performance on the NV split, and detailed task decomposition analyses—are reported in Apps.~\ref{app:additional_results} and ~\ref{app:decomp_analysis}. A comprehensive error analysis is provided in App.~\ref{app:error_analysis}.

\textit{Comparison with CommonsenseQA.} Finally, Table~\ref{tab:csqa_comparison} underscores the central motivation for our benchmark. While LLaMA-3.1-8B achieves 72.2\% accuracy on \textsc{CommonsenseQA}, its performance drops sharply on \LCQA: 72\% on \textsc{AND}, 62.2\% on \textsc{OR}, 42.7\% on \textsc{Mixed}, and only 13.9\% on \textsc{NEITHER/NOR}. This discrepancy shows that benchmarks like \textsc{CommonsenseQA} substantially overestimate models' commonsense reasoning ability by failing to test relational and compositional plausibility judgments.

\begin{table}[t]
\resizebox{\columnwidth}{!}{%
\begin{tabular}{llllll}
\toprule
\textbf{Paradigm}
& \textbf{Model}
& \textbf{AND}
& \textbf{OR}
& \textbf{NN}
& \textbf{MIX} \\
\midrule

\multirow{5}{*}{0-Shot} & LLaMA-3.3-70B
& 80.9 & 70.9 & 13.4 & 53.0 \\

& LLaMA-3.1-8B
& 71.9 & 62.2 & 13.1 & 41.8 \\

& Qwen2.5-7B
& 79.6 & 68.9 & 12.9 & 53.2 \\

& Gemini-2.5-Flash
& 84.1 & 76.7 & 23.5 & 60.9 \\

& Gemini-3-Flash-Preview
& 82.5 & 70.1 & 39.6 & 61.8 \\

\midrule

\multirow{3}{*}{3-Shot} & LLaMA-3.3-70B
& 85.7 & 77.5 & 10.7  & 50.4 \\

& LLaMA-3.1-8B
& 71.3 & 61.3 & 6.1 & 42.8 \\

& Qwen2.5-7B
& 77.9 & 71.9 & 6.8 & 51.4 \\

\midrule

\multirow{3}{*}{Fine-tuned} & Flan-T5-base 
& 92.8 & 92.4 & 89.2 & 89.6 \\

& DeBERTa-v3-base 
& 87.6 & 87.2 & 84.8 & 82.4 \\

& LLaMA-3.1-8B
& 93.6 & 94.4 & 89.5 & 91.5 \\

\bottomrule
\end{tabular}}

\caption{
\textbf{Macro-F1 on the human-validated test set} across all relation types.
NN = \textsc{NEITHER/NOR}, MIX = \textsc{Mixed}
}
\label{tab:main_summary}
\end{table}

\begin{table}[t]
\centering
\small
\begin{tabular}{lll}
\toprule
\textbf{Dataset } & \textbf{Acc} & \textbf{F1} \\
\midrule
\multicolumn{3}{l}{\textbf{\textsc{CommonsenseQA}}} \\
LLaMA-3.1-8B                 & 72.2 & 72.2 \\
\midrule
\multicolumn{3}{l}{\textbf{\LCQA}} \\
\textsc{AND}                 & 72.0 & 71.9 \\
\textsc{OR}                  & 62.2 & 62.2 \\
\textsc{NEITHER/NOR}         & 13.9 & 13.1 \\
\textsc{Mixed}               & 42.7 & 41.8 \\
\bottomrule
\end{tabular}
\caption{0-shot performance of LLaMA-3.1-8B on the original \textsc{CommonsenseQA} and on the human-validated subset of each logical relation in \LCQA.}
\label{tab:csqa_comparison}
\end{table} 
\section{Conclusions and Future Work}
We introduce \LCQA, a benchmark that reframes commonsense reasoning as selecting among logically composed answer options. By combining human-validated atomic statements with symbolic operators that encode conjunction, disjunction, and negation of plausibility, our work exposes reasoning failures obscured by traditional single-answer benchmarks. Experiments reveal that while LLMs handle conjunctive and, to a lesser extent, disjunctive reasoning, they fail sharply on negation-based composition. Additional decomposition analysis suggests that the difficulty arises from the interaction of atomic plausibility estimation, operator grounding, and distractor competition rather than from a single source of error.

Future work includes extending the benchmark to generative settings and structured reasoning frameworks, expanding the operator set, scaling awareness--consensus validation, and studying transfer to open-ended question answering, dialogue, and planning tasks.

\section*{Limitations}

\LCQA\ offers a controlled framework for probing compositional commonsense reasoning, but several limitations remain. The logical operators (AND, OR, and NEITHER/NOR) capture plausibility-level relations rather than full propositional semantics, and do not cover richer structures such as implication, exclusivity, or temporal and causal reasoning. Extending the operator space would enable a more complete evaluation of symbolic and relational inference.

Model coverage is another limitation. Although we evaluate a diverse set of architectures, our study remains limited in scope. Broader model comparisons may further clarify which components most improve logical composition. Further, we do not examine transfer to downstream tasks such as open-ended question answering, dialogue, or planning.
\section*{Ethical Considerations}
\LCQA\ is derived from the publicly available \textsc{CommonsenseQA}
dataset and consists of generic, hypothetical scenarios that do not reference
identifiable individuals or personal contexts. To the best of our knowledge, the
dataset does not contain private or sensitive information.

We used an LLM to initially generate atomic answer options and increase coverage and diversity. These candidates are subsequently refined and partially validated by human annotators using an awareness--consensus framework. As a result, the dataset reflects socially grounded plausibility judgments rather than objective ground truth, and may thus inherit cultural or contextual biases.


\bibliography{custom}

\appendix
\section{Appendix}
\label{sec:appendix}

\subsection{Model Selection}\label{app:models}

This section describes the architectures and evaluation settings of all models considered in our experiments, covering encoder-only, encoder-decoder, and decoder-only LLMs.

\paragraph{Encoder Models}
We fine-tune \textsc{DeBERTa-v3-base}~\cite{he2021debertav3}, a pretrained encoder which encodes the concatenation of the question and one composite option and applies a classification head to score the four candidate answers.

\paragraph{Encoder–Decoder and Multi-Step Reasoning Models}
We also fine-tune sequence-to-sequence models to generate the output label.
\textsc{Flan-T5-base}~\cite{chung2024scaling} is fine-tuned to output a single token corresponding to the chosen option.
Additionally, we evaluate \textsc{Entailer-11B}~\cite{tafjord2022entailer}, a model designed to produce faithful chains-of-reasoning by recursively generating premises through backward-chaining and verifying them via self-querying.  
Although originally developed for faithful explanation generation, we use Entailer as a strong encoder-decoder baseline to examine whether reasoning-oriented models better capture the compositional structures in our dataset.

\paragraph{Decoder-Only LLMs}
We evaluate several open-weight instruction-tuned LLMs in zero- and few-shot settings:
\textsc{Llama-3.3-70B-Instruct},
\textsc{Llama-3.1-8B-Instruct}~\cite{dubey2024llama} and 
\textsc{Qwen2.5-7B-Instruct}~\cite{qwen2025qwen25technicalreport}.
These models are evaluated via prompting and are instructed to output a single capital letter corresponding to the predicted option. In addition to prompted evaluation, \textsc{Llama-3.1-8B-Instruct} is also evaluated under supervised fine-tuning.

We also evaluate proprietary closed-source models in the zero-shot setting, specifically \textsc{Gemini-2.5-Flash}~\cite{comanici2025gemini} and \textsc{Gemini-3-Flash-Preview}. These models are evaluated using the same multiple-choice prompting setup and output constraints as the open-weight decoder-only models.

All models are evaluated on each operator-specific subset defined in Table~\ref{tab:logic_labels}, as well as on the \textsc{Mixed} condition.

\subsection{Prompting Setup}\label{app:prompting}
In this section, we describe the prompting strategy used for decoder-only models. We adopt a unified multiple-choice prompting scheme for both open-weight and proprietary closed-source prompted models.  
Each prompt presents the question, the four composite options, and an instruction to answer with a single letter (A--D). 
We evaluate 0-, 1-, 2- and 3-shot settings using a fixed set of labeled examples. 

Decoding uses \texttt{temperature}$=0.0$, \texttt{top-p}$=0.9$, and \texttt{max\_new\_tokens}$=3$. Outputs that do not correspond to a valid answer option are treated as incorrect.

\paragraph{Chain-of-Thought (CoT)}
We evaluate CoT prompting for \textsc{Llama-3.1-8B-Instruct} only, due to computational cost. CoT runs use a larger generation budget (\texttt{max\_new\_tokens} $= 200$-$300$).

\subsection{Prompt Templates}\label{app:prompts}

This section provides the prompt templates used for dataset construction and evaluation.

\subsubsection{Dataset Construction Prompts}
\label{app:construction_prompts}

We describe the prompts used in each stage of dataset construction below.

\paragraph{Stage 1: Candidate Option Generation}

\begin{quote}\small
Given a commonsense question and its original correct and incorrect answers,
generate additional atomic answer options that require non-trivial reasoning.

\textbf{Instructions:}
\begin{itemize}
    \item Generate 4--6 additional \emph{correct} options that are independently
    plausible and require multi-step commonsense reasoning (e.g., causal,
    situational, or social inference).
    \item Generate 4--6 \emph{incorrect} options that are highly plausible and
    contextually related to the question, but fail due to a subtle violation
    (e.g., incorrect purpose, mismatched context, or missing constraint).
    \item Avoid trivial answers, obvious domain mismatches, or options that can
    be rejected through surface-level keyword matching.
    \item Ensure each option can be evaluated independently, without relying on
    other options or comparisons.
\end{itemize}

\textbf{Output:} A JSON object containing expanded lists of correct and incorrect
atomic answer options.
\end{quote}

\paragraph{Stage 2: Option Refinement and Pruning}

\begin{quote}\small
Given a commonsense question and a set of generated atomic answer options,
refine and filter the options to produce a challenging and internally consistent
reasoning set.

\textbf{Instructions:}
\begin{itemize}
    \item Remove options that are logically inconsistent, factually incorrect,
    trivial, or mismatched to the question category.
    \item Refine correct options to eliminate direct or overly obvious answers
    while preserving plausibility and the need for multi-step inference.
    \item Refine incorrect options into near-miss distractors that satisfy most
    semantic and contextual constraints but fail on exactly one subtle causal,
    temporal, or pragmatic nuance.
    \item Enforce consistency in option length and prohibit explanatory or
    justificatory phrasing.
\end{itemize}

\textbf{Output:} A JSON object containing a fixed set of refined correct and incorrect atomic options.
\end{quote}

\paragraph{Stage 3: Logical Composition}
In Stage~3, refined atomic options are combined using deterministic symbolic
rules to form composed answer options labeled with one of the three base
plausibility operators: \textsc{AND}, \textsc{OR}, and \textsc{NEITHER/NOR}.
We additionally construct a \textsc{Mixed} setting in which different operators
appear across answer choices within the same question. This stage does not
involve prompting or any additional model inference.

\subsubsection{Evaluation Prompts}
\label{app:evaluation_prompts}
Prompts differ only in the number of in-context examples and the logical operator expressed in the answer options.

\paragraph{Zero-shot Prompt}
In the zero-shot setting, we do not provide any labeled examples.

\begin{quote}\small
Answer the following commonsense question by selecting the correct option.
Respond with \textbf{only} the single capital letter (A, B, C, or D).

Question: <QUESTION>

A. <OPTION A>  \\
B. <OPTION B>  \\
C. <OPTION C>  \\
D. <OPTION D>  \\

Answer:
\end{quote}

\paragraph{Few-shot Prompting}
For few-shot (1-, 2-, and 3-shot) evaluation, we prepend labeled examples before the target question. \textbf{The 1- and 2-shot prompts are strict prefixes of the 3-shot prompt},
using the first one or two examples respectively. 

For operator-specific subsets (\textsc{AND}, \textsc{OR}, \textsc{NEITHER/NOR}),
all in-context examples match the operator being evaluated. For the
\textsc{Mixed} setting, examples are sampled across operators.

\paragraph{Conjunctive Reasoning (\textsc{AND}) Prompt}
\begin{quote}\small
Question: Sammy wanted to go to where the people were. Where might he go?

A. social venues AND quiet retreats \\
B. local events AND social venues \\ 
C. sports arenas AND quiet retreats \\
D. sports arenas AND train platforms  \\

Answer: B

\vspace{0.6em}

Question: The fox walked from the city into the forest, what was it looking for?

A. suitable prey AND shelter from disturbances  \\
B. urban garden AND farm fields  \\
C. natural habitat AND farm fields \\ 
D. suitable prey AND neighborhood pets  \\

Answer: A

\vspace{0.6em}

Question: What home entertainment equipment requires cable?

A. wireless speaker AND portable projector \\
B. television AND home theater system \\
C. wireless speaker AND gaming console \\
D. digital frame AND portable projector \\

Answer: B

\vspace{0.6em}
Now answer the following commonsense question by selecting the correct option.
Respond with \textbf{only} the single capital letter (A, B, C, or D).

Question: <TARGET AND QUESTION>

A. <OPTION A>  \\
B. <OPTION B>  \\
C. <OPTION C>  \\
D. <OPTION D>  \\

Answer:

\end{quote}

\paragraph{Disjunctive Reasoning (\textsc{OR}) Prompt}
\begin{quote}\small
Question: Sammy wanted to go to where the people were. Where might he go?

A. quiet retreats OR empty parks  \\
B. local events OR empty parks  \\
C. sports arenas OR empty parks  \\
D. train platforms OR empty parks  \\

Answer: B

\vspace{0.6em}

Question: The forgotten leftovers had gotten quite old; he found it covered in
mold in the back of his what?

A. living room OR dining area  \\
B. utility drawer OR kitchen counter \\ 
C. living room OR utility drawer  \\
D. cold storage OR utility drawer \\ 

Answer: D

\vspace{0.6em}

 Question: What do people use to absorb extra ink from a fountain pen? 

A. absorbent paper OR notebook cover \\
B. ink storage OR writing surface \\ 
C. notebook cover OR writing surface \\
D. ink storage OR fabric swatch \\

Answer: A

\vspace{0.6em}
Now answer the following commonsense question by selecting the correct option.
Respond with \textbf{only} the single capital letter (A, B, C, or D).

Question:  <TARGET OR QUESTION>

A. <OPTION A>  \\
B. <OPTION B>  \\
C. <OPTION C>  \\
D. <OPTION D>  \\

Answer:
\end{quote}

\paragraph{Negation-Based Reasoning (\textsc{NEITHER/NOR}) Prompt}
\begin{quote}\small
Question: Sammy wanted to go to where the people were. Where might he go?

A. NEITHER local events NOR train platforms  \\
B. NEITHER sports arenas NOR train platforms  \\
C. NEITHER social venues NOR sports arenas  \\
D. NEITHER social venues NOR quiet retreats  \\

Answer: B

\vspace{0.6em}

Question: The only baggage the woman checked was a drawstring bag. Where was she
heading with it?

A. NEITHER family gathering NOR local gathering  \\
B. NEITHER airport travel NOR local gathering  \\
C. NEITHER local gathering NOR short trip  \\
D. NEITHER business engagement NOR local gathering  \\

Answer: C

\vspace{0.6em}

Question: The forgotten leftovers had gotten quite old, he found it covered in mold in the back of his what? 

A. NEITHER cold storage NOR dining area \\
B. NEITHER sealed container NOR living room \\
C. NEITHER food cabinet NOR utility drawer \\
D. NEITHER living room NOR utility drawer \\

Answer: D

\vspace{0.6em}
Now answer the following commonsense question by selecting the correct option.
Respond with \textbf{only} the single capital letter (A, B, C, or D).

Question: <TARGET NEITHER/NOR QUESTION>

A. <OPTION A>  \\
B. <OPTION B>  \\
C. <OPTION C>  \\
D. <OPTION D>  \\

Answer:

\end{quote}

\paragraph{Mixed Prompt}
\begin{quote}\small
Question: The fox walked from the city into the forest, what was it looking for?

A. natural habitat AND shelter from disturbances \\
B. farm fields OR neighborhood pets \\
C. city park AND neighborhood pets \\
D. farm fields AND neighborhood pets \\

Answer: A 

\vspace{0.6em}

Question: Google Maps and other highway and street GPS services have replaced what? 

A. historical navigation logs AND tourist information centers \\
B. NEITHER printed road maps NOR route planning software \\
C. manual navigation techniques AND tourist information centers \\
D. manual navigation techniques AND route planning software \\

Answer: D 

\vspace{0.6em}

Question: Where is a business restaurant likely to be located? 

A. corporate office clusters AND at university campuses \\
B. business park entrances AND near conference venues \\
C. NEITHER near conference venues NOR at university campuses \\
D. NEITHER near conference venues NOR near tourist sites \\

Answer: B

\vspace{0.6em}
Now answer the following commonsense question by selecting the correct option.
Respond with \textbf{only} the single capital letter (A, B, C, or D).

Question: <TARGET MIXED QUESTION>

A. <OPTION A>  \\
B. <OPTION B>  \\
C. <OPTION C>  \\
D. <OPTION D>  \\

Answer:
\end{quote}

\paragraph{Chain-of-Thought Prompting}
For CoT evaluation, we add a brief instruction asking the
model to reason about the plausibility of each atomic statement and how the
logical operator applies before providing the final answer.

\begin{quote}\small
Carefully reason about each atomic statement and the logical operator.
Then provide the final answer as a single capital letter (A--D).
\end{quote}

\paragraph{Decoding Constraints}
All decoder-only models are evaluated with near-deterministic decoding
(\texttt{temperature}=0.0, \texttt{top-p}=0.9).
Outputs are constrained to a single capital letter (A--D); any other output is
treated as incorrect.

\subsection{Annotation Guidelines}
\label{app:annotation_guidelines}

This section describes the guidelines provided to human annotators for validating
refined atomic answer options in Stage~2 of dataset construction.

\paragraph{Annotation Task}
Each annotation item consists of a commonsense question and a set of
automatically labeled answer options. Options are pre-labeled as \emph{correct}
or \emph{incorrect} by the refinement pipeline. Annotators are instructed to
evaluate each option \emph{independently}, without reconsidering or changing the
question itself.

The goal of annotation is only \emph{verification}: annotators judge
whether the assigned labels are reasonable under commonsense reasoning.

\paragraph{Evaluation Criteria}
For each option, annotators provide judgments along two dimensions, following the
awareness--consensus framework of \citet{whiting2024commonsense}:

\begin{enumerate}
    \item \textbf{Correctness (Awareness).}
    Annotators indicate whether they personally believe the option is a valid
    answer to the question.
    \begin{itemize}
        \item For options labeled as \emph{correct}: “Do you believe this option
        is actually a valid answer?”
        \item For options labeled as \emph{incorrect}: “Do you believe this option
        is actually wrong or implausible?”
    \end{itemize}

    \item \textbf{Commonsense (Consensus).}
    Annotators indicate whether they believe most adults would agree with the
    judgment using everyday knowledge.
\end{enumerate}

Each criterion is rated using a three-point scale: \texttt{Yes}, \texttt{No}, or
\texttt{Maybe}. Annotators are required to provide a brief comment for any
\texttt{No} or \texttt{Maybe} response (e.g., “could be true in some contexts” or
“requires specialized knowledge”).

\paragraph{Definition of Commonsense}
Annotators are instructed to interpret commonsense as knowledge that an average
adult could reasonably be expected to possess, without relying on formal
education, technical expertise, or domain-specific facts. Options requiring
specialized scientific, historical, or technical knowledge are marked
accordingly.

\paragraph{Handling Ambiguity}
Annotators are encouraged to consider realistic context and pragmatic reasoning.
When an option is plausible only under rare or contrived interpretations, it
should be marked as \texttt{Maybe} rather than \texttt{Yes}. The goal is to assess
typical human judgments, not theoretical possibility.

\paragraph{Adjudication}
Two annotators independently evaluate each item. Disagreements, as well as any
option receiving a \texttt{Maybe} are resolved
through adjudication by a third reviewer. Only disputed cases are adjudicated;
non-disputed options retain their original labels.

\paragraph{Illustrative Examples}
Annotators are provided with worked examples illustrating borderline cases, such
as partially plausible distractors, context-dependent interpretations, and
options that are topically related but logically invalid. These examples are used
to clarify expectations but are not treated as templates.

\subsection{Training Details}\label{app:training}
This section reports the training configurations used across all evaluated models, as well as the hardware used in our experiments. Prompting and decoding details for decoder-only models are described separately in Appendix~\ref{app:prompting}.

Encoder-only and encoder--decoder models are fine-tuned with AdamW
($5\times10^{-5}$ learning rate, weight decay $0.01$) for up to 10 epochs with early stopping (patience 3), using a maximum sequence length of 512 and an effective batch size of 8.  

For supervised fine-tuning of \textsc{Llama-3.1-8B-Instruct}, we use QLoRA with 4-bit NF4 quantization and LoRA adapters (rank 16, $\alpha=32$, dropout 0.05). The model is fine-tuned with AdamW ($2\times10^{-4}$ learning rate, weight decay $0.01$) for up to 10 epochs with early stopping (patience 3), using a maximum sequence length of 512 and an effective batch size of 16.

Most experiments run on NVIDIA P100 GPUs. Only the largest models, CoT
evaluations, and decoder-only fine-tuning runs are conducted on A100 GPUs.

\subsection{Additional Results}\label{app:additional_results}

This section reports supplementary experimental results that complement the main findings discussed in Section~\ref{sec:experiments}.
Table~\ref{tab:main_results_app} reports accuracy and macro-F1 across all logical relations for human-validated (HV) and non-human-validated (NV) test subsets, including instruction-tuned and fine-tuned models.  Table~\ref{tab:shots_decoder_models_hv_app} presents macro-F1 for decoder-only LLMs under 0--3 shot prompting and CoT settings, including zero-shot results for proprietary closed-source models. Encoder-only and encoder--decoder models without fine-tuning exhibit uniformly low performance. Overall trends mirror the main paper: reasonable performance on \textsc{AND} and \textsc{OR}, and collapse on \textsc{NEITHER/NOR} across settings.

\begin{table*}[t]
\centering
\scriptsize
\renewcommand{\arraystretch}{0.9}
\setlength{\tabcolsep}{3pt}

\begin{tabularx}{\textwidth}{l *{16}{>{\centering\arraybackslash}X}}
\toprule
\multirow{2}{*}{\textbf{Model}}
& \multicolumn{4}{c}{\textbf{AND}}
& \multicolumn{4}{c}{\textbf{OR}}
& \multicolumn{4}{c}{\textbf{NN}}
& \multicolumn{4}{c}{\textbf{MIX}} \\
\cmidrule(lr){2-5}
\cmidrule(lr){6-9}
\cmidrule(lr){10-13}
\cmidrule(lr){14-17}
& NV A & NV F1 & HV A & HV F1
& NV A & NV F1 & HV A & HV F1
& NV A & NV F1 & HV A & HV F1
& NV A & NV F1 & HV A & HV F1 \\
\midrule

Gemini-2.5-Flash
& 80.5 & 80.5 & 84.0 & 84.1
& 70.0 & 70.0 & 76.2 & 76.7
& 25.1 & 24.2 & 23.9 & 23.5
& 55.3 & 55.3 & 60.9 & 60.9 \\

Gemini-3-Flash-Preview
& 83.0 & 83.1 & 82.5 & 82.5
& 69.8 & 69.9 & 69.8 & 70.1
& 34.3 & 32.9 & 39.9 & 39.6
& 55.1 & 54.7 & 62.0 & 61.8 \\

LLaMA-70B
& 76.4 & 76.2 & 80.8 & 80.9
& 72.4 & 72.3 & 71.2 & 70.9
& 13.6 & 13.4 & 14.0 & 13.4
& 53.7 & 53.4 & 53.2 & 53.0 \\

LLaMA-8B
& 67.2 & 66.7 & 72.0 & 71.9
& 63.2 & 62.6 & 62.2 & 62.2
& 8.6 & 7.9 & 13.9 & 13.1
& 44.5 & 43.9 & 42.7 & 41.8 \\

Qwen2.5-7B
& 76.2 & 76.2 & 79.6 & 79.6
& 70.4 & 70.5 & 68.8 & 68.9
& 13.6 & 11.8 & 13.3 & 12.9
& 54.0 & 53.7 & 53.2 & 53.2 \\

\midrule

DeBERTa-v3-base
& 28.8 & 28.6 & 27.2 & 27.2
& 23.6 & 23.8 & 23.2 & 23.1
& 30.8 & 30.9 & 27.2 & 27.4
& 28.8 & 28.8 & 28.4 & 28.4 \\

DeBERTa-v3-base (FT)
& 90.8 & 90.8 & 87.6 & 87.6
& 88.0 & 88.0 & 87.2 & 87.2
& 89.2 & 89.2 & 84.8 & 84.8
& 84.8 & 84.8 & 82.4 & 82.4 \\

Entailer-11B
&  5.6 &  5.5 & 3.2 & 3.2
& 20.8 & 20.8 & 18.4 & 18.4
& 22.0 & 22.0 & 24.0 & 24.1
& 16.8 & 16.8 & 22.4 & 22.3 \\

Flan-T5-base
& 38.8 & 38.0 & 37.2 & 36.2
& 63.2 & 62.2 & 65.2 & 64.3
& 12.0 & 11.4 & 16.8 & 16.3
& 32.8 & 32.5 & 33.6 & 33.0 \\

Flan-T5-base (FT)
& 94.8 & 94.8 & 92.8 & 92.8
& 97.6 & 97.6 & 92.4 & 92.4
& 93.2 & 93.2 & 89.2 & 89.2
& 87.2 & 87.2 & 89.6 & 89.6 \\

LLaMA-3.1-8B (FT)
& 96.0 & 96.0 & 93.6 & 93.6
& 97.2 & 97.2 & 94.4 & 94.4
& 93.2 & 93.2 & 89.6 & 89.5
& 91.6 & 91.6 & 91.6 & 91.5 \\

\bottomrule
\end{tabularx}

\caption{
\textbf{0-shot vs.\ fine-tuned:}
Accuracy (A) and macro-F1 across all logical relations.
NV = non-human-validated labels; HV = human-validated subset.
NN = \textsc{NEITHER/NOR}, MIX = \textsc{Mixed}, FT = fine-tuned.
}
\label{tab:main_results_app}
\end{table*}

\begin{table}[t]
\centering
\scriptsize
\renewcommand{\arraystretch}{0.85}
\setlength{\tabcolsep}{2.5pt}
\resizebox{\columnwidth}{!}{%
\begin{tabular}{lcccccccc}
\toprule
\multirow{2}{*}{\textbf{Model}}
& \multicolumn{2}{c}{\textbf{AND}}
& \multicolumn{2}{c}{\textbf{OR}}
& \multicolumn{2}{c}{\textbf{NN}}
& \multicolumn{2}{c}{\textbf{MIX}} \\
\cmidrule(lr){2-3}
\cmidrule(lr){4-5}
\cmidrule(lr){6-7}
\cmidrule(lr){8-9}
& NV F1 & HV F1
& NV F1 & HV F1
& NV F1 & HV F1
& NV F1 & HV F1 \\
\midrule

\multicolumn{9}{l}{\textbf{0-shot}} \\
LLaMA-70B & 76.2 & 80.9 & 72.3 & 70.9 & 13.4 & 13.4 & 53.4 & 53.0 \\
LLaMA-8B  & 66.7 & 71.9 & 62.6 & 62.2 & 7.9 & 13.1 & 43.9 & 41.8 \\
Qwen-7B   & 76.2 & 79.6 & 70.5 & 68.9 & 11.8 & 12.9 & 53.7 & 53.2 \\
Gemini-2.5-Flash         & 80.5 & 84.1 & 70.0 & 76.7 & 24.2 & 23.5 & 55.3 & 60.9 \\
Gemini-3-Flash-Preview   & 83.1 & 82.5 & 69.9 & 70.1 & 32.9 & 39.6 & 54.7 & 61.8 \\
\midrule
\multicolumn{9}{l}{\textbf{1-shot}} \\
LLaMA-70B & 80.3 & 84.8 & 76.9 & 79.5 & 7.9 & 11.3 & 54.7 & 53.1 \\
LLaMA-8B  & 72.1 & 67.0 & 61.7 & 62.0 & 5.9 & 5.3 & 45.0 & 41.7 \\
Qwen-7B   & 73.9 & 79.6 & 75.0 & 71.6 & 5.7 & 7.4 & 57.4 & 55.4 \\

\midrule
\multicolumn{9}{l}{\textbf{2-shot}} \\
LLaMA-70B & 78.7 & 83.6 & 74.0 & 76.8 & 9.1 & 12.2 & 53.1 & 51.7 \\
LLaMA-8B  & 71.4 & 75.4 & 61.6 & 63.9 & 6.0 & 6.6 & 46.1 & 41.2 \\
Qwen-7B   & 72.3 & 77.5 & 70.6 & 69.6 & 5.7 & 6.9 & 55.2 & 52.0 \\

\midrule
\multicolumn{9}{l}{\textbf{3-shot}} \\
LLaMA-70B & 80.0 & 85.7 & 75.1 & 77.5 & 7.8 & 10.7 & 52.4 & 50.4 \\
LLaMA-8B  & 75.8 & 71.3 & 61.0 & 61.3 & 5.5 & 6.1 & 47.0 & 42.8 \\
Qwen-7B   & 73.4 & 77.9 & 73.2 & 71.9 & 4.6 & 6.8 & 55.5 & 51.4 \\

\midrule
\multicolumn{9}{l}{\textbf{CoT (0-shot)}} \\
LLaMA-8B  & 65.3 & 67.9 & 63.0 & 62.5 & 10.7 & 8.2 & 45.8 & 42.7 \\

\bottomrule
\end{tabular}
}

\caption{
Macro-F1 across all logical relations under 0--3 shot prompting and chain-of-thought (CoT).
NV = non-human-validated labels; HV = human-validated subset.
NN = \textsc{NEITHER/NOR}, MIX = \textsc{Mixed}.
}
\label{tab:shots_decoder_models_hv_app}
\end{table}

\subsection{Task Decomposition Analysis}\label{app:decomp_analysis}

Although \LCQA\ can be conceptually decomposed into atomic plausibility estimation followed by logical composition, this decomposition does not fully explain the observed task difficulty in the full multiple-choice setting. To examine this, we separate atomic plausibility estimation, operator grounding, and distractor competition, and evaluate each component independently using \textsc{Llama-3.1-8B}.

\begin{table}[t]
\centering
\scriptsize
\renewcommand{\arraystretch}{0.85}
\begin{tabular}{lcc}
\toprule
\textbf{Metric} & \textbf{HV} & \textbf{NV} \\
\midrule
Accuracy & 79.4 & 78.5 \\
Macro-F1 & 77.8 & 78.1 \\
\bottomrule
\end{tabular}
\caption{Atomic plausibility classification on independently evaluated Stage~2 atomic statements, without logical composition.}
\label{tab:atomic_plausibility_app}
\end{table}

As shown in Table~\ref{tab:atomic_plausibility_app}, the model achieves reasonably strong performance when classifying atomic statements independently, indicating substantial atomic commonsense knowledge. The remaining question is whether the difficulty of \LCQA\ arises primarily from operator grounding or from the interaction between atomic plausibility errors and competitive multiple-choice composition.

\begin{table*}[t]
\centering
\scriptsize
\begin{tabular}{lcccccccc}
\toprule
\textbf{Setting} & \textbf{AND (HV)} & \textbf{AND (NV)} & \textbf{OR (HV)} & \textbf{OR (NV)} & \textbf{NN (HV)} & \textbf{NN (NV)} & \textbf{MIX (HV)} & \textbf{MIX (NV)} \\
\midrule
Composite Task & 72.0 & 67.2 & 62.2 & 63.2 & 13.9 & 8.6 & 42.7 & 44.5 \\
Operator Verification & 69.2 & 64.8 & 51.6 & 51.6 & 58.8 & 61.2 & 61.3 & 57.3 \\
Atomic + Compose (No Distractors) & 74.8 & 69.2 & 90.8 & 90.4 & 71.2 & 78.4 & 78.3 & 76.5 \\
Atomic + Compose (With Distractors) & 68.4 & 63.6 & 95.6 & 97.2 & 46.8 & 48.0 & 65.1 & 63.7 \\
\bottomrule
\end{tabular}
\caption{Decomposition analysis of \textsc{Llama-3.1-8B}, separating atomic plausibility estimation, operator grounding, and distractor effects.}
\label{tab:decomposition_analysis_app}
\end{table*}

Table~\ref{tab:decomposition_analysis_app} clarifies this decomposition. In the Operator Verification setting, the model is given gold atomic plausibility labels and must determine whether they satisfy the target relation. Performance remains imperfect across all relation types, showing that logical composition is itself non-trivial even when plausibility estimation is removed. In contrast, Atomic + Compose (No Distractors) performs relatively strongly, especially for \textsc{OR}, suggesting that atomic plausibility predictions combined with deterministic composition can recover many correct answers when no competing options are present.
However, performance changes once distractor composites are introduced. Under Atomic + Compose (With Distractors), performance drops substantially, indicating that the full multiple-choice setting imposes additional difficulty beyond atomic classification and rule application alone. This effect is especially pronounced for \textsc{NEITHER/NOR}, which degrades sharply under distractor competition, whereas \textsc{OR} remains comparatively strong. Taken together, these results suggest that the difficulty of \LCQA\ does not arise solely from atomic plausibility estimation, operator grounding, or distractor competition, but from their interaction in the full multiple-choice setting.

\subsection{Error Analysis}\label{app:error_analysis}

We analyze errors made by \textsc{LLaMA-3.1-8B} in the zero-shot setting. Table~\ref{tab:error_examples_app} summarizes representative failure patterns revealed by \LCQA.~Because each option is a compositional pair of atomic statements, errors expose not only mis-classifications but also deeper limitations highlighting how models fail to compose plausibility across logical operators.

\begin{table}[t]
\centering
\scriptsize
\begin{tabular}{p{0.9cm} p{1.9cm} p{4.0cm}}
\toprule
\textbf{Op} & \textbf{Failure Pattern} & \textbf{Representative Error (Gold-Pred)} \\
\midrule

\textbf{AND} 
& Single-statement dominance 
& \textbf{Q:} Heat applied to combustible materials? \newline
Gold: emits toxic fumes AND decomposes into gases \newline
Pred: ignites rapidly AND melts into liquid \\
\midrule

\textbf{OR}
& \makecell[l]{Plausibility\\collapse}
& \textbf{Q:} Heat applied to combustible materials? \newline
Gold: ignites rapidly OR melts into liquid \newline
Pred: melts into liquid OR burns without oxygen \\
\midrule

\textbf{NEITHER} 
& Negation inversion 
& \textbf{Q:} Where do you buy a glass of wine? \newline
Gold: NEITHER public park NOR home kitchen  \newline
Pred: NEITHER licensed restaurant NOR home kitchen \\
\cmidrule(lr){2-3}

& \makecell[l]{Plausibility\\dominance}
& \textbf{Q:} Where might a child with no home go? \newline
Gold: NEITHER school event NOR family gathering \newline
Pred: NEITHER youth shelter NOR foster care \\
\bottomrule
\end{tabular}
\caption{Representative error patterns for \textsc{LLaMA-3.1-8B} (0-shot) across \textsc{AND}, \textsc{OR}, and \textsc{NEITHER/NOR}.}
\label{tab:error_examples_app}
\end{table}

\paragraph{Conjunctive and Disjunctive Reasoning}
Across both AND and OR settings, the dominant failure mode is the model's
tendency to recognize only a \emph{single} plausible statement and ignore how
plausibility must be composed across clauses.
In AND questions, this appears as \emph{single-statement dominance}: the model anchors on a highly plausible clause and fails to reject an implausible partner. For example, in the combustibility question, it predicts \emph{ignites rapidly AND melts into liquid}, correctly identifying ignition
while overlooking that combustible materials do not melt. 

The same underlying behavior manifests in OR questions. Rather than exploiting the requirement that at least one clause be plausible, the model treats OR as indicating thematic similarity. In the combustibility example, it predicts
\emph{melts into liquid OR burns without oxygen}, selecting two implausible but fire-related statements. These errors indicate that models do not enforce the compositional constraint
imposed by the operator; instead, they rely on surface-level plausibility cues
and semantic similarity.

\paragraph{Negation-Based Reasoning}
NEITHER/NOR proves most challenging.  
We observe \emph{negation inversion}, where the model selects the 
\emph{most plausible} pair of statements despite the operator requiring both 
to be implausible.  
For example, in the wine-purchase question, the model predicts  
\emph{NEITHER licensed restaurant NOR home kitchen}, incorrectly negating the 
true location.  
A second pattern, \emph{plausibility dominance}, reflects the model's tendency 
to retain highly plausible statements even under negation. For example, in the 
homeless-child example, the model selects  
\emph{NEITHER youth shelter NOR foster care}, two locations that are plausible options.  
These behaviors indicate a broader difficulty with composing negation and 
plausibility simultaneously.

\end{document}